\documentclass[conference]{IEEEtran}
\IEEEoverridecommandlockouts
\usepackage{cite}
\usepackage{amsmath,amssymb,amsfonts}
\usepackage{algorithmic}
\usepackage{graphicx}
\usepackage{comment}
\usepackage{textcomp}
\def\BibTeX{{\rm B\kern-.05em{\sc i\kern-.025em b}\kern-.08em
    T\kern-.1667em\lower.7ex\hbox{E}\kern-.125emX}}
\usepackage{xcolor}
\usepackage{adjustbox} 
\usepackage{tabularx} 
\usepackage{CJKutf8}
\usepackage{xcolor}
\usepackage{pict2e}
\usepackage{color}
\usepackage{multirow}
\usepackage{xcolor}
\usepackage{pict2e}
\usepackage{color}
\usepackage{multirow}
\usepackage{url}
\def\BibTeX{{\rm B\kern-.05em{\sc i\kern-.025em b}\kern-.08em
    T\kern-.1667em\lower.7ex\hbox{E}\kern-.125emX}}
\begin{document}

\title{Enhancing PTSD Outcome Prediction with Ensemble Models in Disaster Contexts}


\author{\IEEEauthorblockN{1\textsuperscript{st}Ayesha Siddiqua}
\IEEEauthorblockA{\textit{Department of Computer} \\ \textit{Science and Engineering} \\
\textit{Northern University Bangladesh}\\
Dhaka, Bangladesh \\
ayesha.siddiqua@nub.ac.bd}
\and
\IEEEauthorblockN{2\textsuperscript{nd}Atib Mohammad Oni}
\IEEEauthorblockA{\textit{Department of Electrical and}\\ 
\textit{Brac University}\\
Dhaka, Bangladesh \\
atib.mohammad@bracu.ac.bd}
\and
\IEEEauthorblockN{ 3\textsuperscript{rd} Abu Saleh Musa Miah, \\*4\textsuperscript{rd} Jungpil Shin}
\IEEEauthorblockA{\textit{School of Computer Science and Engineering } \\
\textit{The University of Aizu}\\
Aizuwakamatsu, Fukushima 965-8580, Japan\\
musa@u-aizu.ac.jp,jpshin@u-aizu.ac.jp}
}

\maketitle

\begin{abstract}
Post-traumatic stress disorder (PTSD) is a significant mental health challenge that affects individuals exposed to traumatic events. Early detection and effective intervention for PTSD are crucial, as it can lead to long-term psychological distress if untreated. Accurate detection of PTSD is essential for timely and targeted mental health interventions, especially in disaster-affected populations.
Existing research has explored machine learning approaches for classifying PTSD, but many face limitations in terms of model performance and generalizability. To address these issues, we implemented a comprehensive preprocessing pipeline. This included data cleaning, missing value treatment using the SimpleImputer, label encoding of categorical variables, data augmentation using SMOTE to balance the dataset, and feature scaling with StandardScaler. The dataset was split into 80\% training and 20\% testing. We developed an ensemble model using a majority voting technique among several classifiers, including Logistic Regression, Support Vector Machines (SVM), Random Forest, XGBoost, LightGBM, and a customized Artificial Neural Network (ANN). The ensemble model achieved an accuracy of 96.76\% with a benchmark dataset, significantly outperforming individual models. The proposed method's advantages include improved robustness through the combination of multiple models, enhanced ability to generalize across diverse data points, and increased accuracy in detecting PTSD. Additionally, the use of SMOTE for data augmentation ensured better handling of imbalanced datasets, leading to more reliable predictions. The proposed approach offers valuable insights for policymakers and healthcare providers by leveraging predictive analytics to address mental health issues in vulnerable populations, particularly those affected by disasters.
\end{abstract}


\section{Introduction}
\label{sec: Introduction}
Post-Traumatic Stress Disorder (PTSD) is a debilitating mental health condition caused by exposure to a traumatic event, such as military warfare, natural disasters, accidents, or assaults. Symptoms of PTSD, including intrusive thoughts, flashbacks, nightmares, and acute anxiety, can significantly affect the quality of life and daily functioning.  Among the mental health conditions that may arise following exposure to or observation of traumatic incidents, Post-Traumatic Stress Disorder (PTSD) is particularly critical\cite{yehuda2015post}. PTSD is a substantial concern, as it negatively impacts an individual’s ability to function effectively in everyday life, affecting interpersonal relationships, occupational performance, and overall mental well-being. Natural disasters have far-reaching effects on mental health, often leading to significant psychological disorders such as Post-Traumatic Stress Disorder (PTSD) \cite{mao2021role}\cite{seddiky2024climate} \cite{heanoy2024impact}. Individuals who have encountered traumatic experiences, such as earthquakes, floods, and hurricanes, frequently report increased levels of anxiety, depression, and stress-related symptoms \cite{mao2021role}\cite{yehuda2015post}.\\
Situated in a disaster-prone South Asian region, Bangladesh is exceptionally vulnerable to various natural catastrophes, including cyclones, floods, and riverbank erosion. The country's geographical context, characterized by low-lying terrain and extensive river networks, exacerbates the hazards associated with these disasters\cite{bhowm21}\cite{rakib2019investigation,hossain2023stochastic}. Given the frequency and severity of such occurrences, understanding and addressing the mental health needs of individuals in Bangladesh is of utmost importance. The urgent need to address mental health repercussions in disaster-affected populations has spurred researchers and practitioners to explore innovative methods for assessment and intervention \cite{lopez22}. \\
This study conducts a comparative analysis of a customized Artificial Neural Network (ANN) model against various traditional and advanced machine learning algorithms, aiming to determine the most effective method for PTSD classification using survey data. Machine learning (ML) has emerged as a transformative methodology in mental health research, enabling the analysis of complex datasets to uncover patterns and predictors of psychological disorders \cite{zandvakili2020mapping}. By employing various algorithms, researchers can identify risk factors associated with PTSD, providing valuable insights that inform preventive strategies and therapeutic interventions. Traditional ML algorithms, such as Logistic Regression, Support Vector Machines (SVM)\cite{kessler2017trauma,miah2021alzheimer,kafi2022lite_kidney_miah}, Random Forest, XGBoost, and LightGBM,\cite{miah2022natural_EEG,miah2021event_EEG,ge2020identifying} have demonstrated effectiveness in classification tasks across numerous domains, allowing for the identification of key predictors for PTSD \cite{han2023predicting} \cite{bari2023potential}.\\
In addition to these traditional methods, ANNs have gained prominence due to their ability to model complex relationships within data \cite{miah2021alzheimer,miah2023skeleton_euvip,miah2023dynamic_graph_general,miah2023dynamic_mcsoc}. ANNs can capture non-linear patterns, making them particularly suitable for scenarios where relationships between variables are intricate and multifaceted \cite{neria2008post,muntaqim2024eye_miah}. The flexibility of ANN architectures allows for customization to meet the specific needs of the dataset, thereby enhancing predictive performance \cite{razavi2024machine}. \\
This study aims to explore the prevalence of post-traumatic stress disorder (PTSD) in disaster-prone populations by employing a range of machine learning models, including traditional algorithms and a customized Artificial Neural Network (ANN). By comparing the performance of these models, the research seeks to identify the most effective approach for predicting PTSD risk, ultimately contributing to a deeper understanding of mental health outcomes in individuals affected by natural disasters. The findings pave the way for targeted interventions and improved support systems in vulnerable populations.
To enhance model accuracy and reliability, we developed an ensemble model using a majority voting technique, combining Logistic Regression, Support Vector Machines (SVM), Random Forest, XGBoost, LightGBM, and a customized ANN. This ensemble approach achieved a remarkable accuracy of 96.76\% on a benchmark dataset, significantly outperforming individual models. 
Our contributions include:

\begin{itemize}
    \item \textbf{Ensemble Model for PTSD Detection}: The ensemble model offers improved robustness by leveraging the strengths of multiple classifiers, resulting in enhanced predictive performance.
    
    \item \textbf{Comprehensive Preprocessing Pipeline}: We introduced a robust preprocessing framework that includes data cleaning, missing value treatment with SimpleImputer, label encoding of categorical features, and Synthetic Minority Oversampling Technique (SMOTE) for data augmentation, addressing issues with imbalanced datasets and enhancing the model's generalizability.
    
    \item \textbf{Application to Real-World PTSD Data}: The proposed method demonstrates practical application in real-world PTSD survey data, offering actionable insights for disaster-related mental health interventions.
    
    \item \textbf{Predictive Analytics for Mental Health}: Our study showcases the potential of predictive analytics to aid policymakers and healthcare providers in identifying at-risk individuals, enabling the design of effective, targeted mental health interventions in disaster-affected populations.
\end{itemize}

\section{Literature Review}
\label{sec: Literature Review}
Research has consistently shown that exposure to natural disasters is associated with elevated rates of PTSD, anxiety, and depression. A comprehensive meta-analysis by \cite{razavi2024machine} highlighted the prevalence of PTSD and other mental health disorders in populations affected by various disasters, emphasizing the need for timely psychological interventions.\\
The application of machine learning techniques \cite{miah2021alzheimer,kafi2022lite_kidney_miah,10624624_lstm_najmul_miah_har_conference} in mental health research has proven beneficial in identifying patterns and predictors of PTSD. For example, a study by Kessler et al. \cite{kessler2017trauma} utilized machine learning methods to analyze survey data and identified demographic and environmental factors that predict PTSD symptoms. \\
In recent years, more complex models, including ensemble methods like XGBoost and LightGBM, have been leveraged to improve predictive accuracy in mental health assessments. Research
by Ge et al. \cite{ge2020identifying} used the XGBoost machine learning model to predict posttraumatic stress disorder (PTSD) in 2,099 young earthquake survivors. Various combinations of risk factors were evaluated, achieving prediction accuracies between 66\% and 80\%.\\
Schultebraucks et al. \cite{schultebraucks2022deep} explored the use of machine learning-based computer vision and acoustic analysis to classify major depressive disorder (MDD) and posttraumatic stress disorder (PTSD) from free speech responses of 81 patients one month post-trauma. Banerjee et al. \cite{banerjee2019deep} explore the use of speech signal analysis for diagnosing post-traumatic stress disorder (PTSD), addressing limitations of traditional structured interviews. A deep belief network (DBN) model, enhanced by transfer learning from the TIMIT database, was developed to analyze speech features. Results showed improved accuracy in PTSD detection, with the DBN achieving 74.99\%, compared to 57.68\% with the support vector machine classifier.\\
In Bangladesh, the application of ML in mental health research is still in its nascent stages, yet preliminary studies indicate a promising direction. A recent study by Siddik et al. \cite{siddik2024climate} examined the mental health of adolescents in flood shelters in northeastern Bangladesh following the 2022 floods. Findings revealed high rates of mental health issues, with 60.72\% of boys and 80\% of girls experiencing post-traumatic stress disorder (PTSD). The findings support the notion that tailored machine-learning approaches can significantly enhance our understanding of mental health outcomes in disaster-prone populations.   \\

\section{Proposed Methodology}
\label{sec: Proposed Methodology}
\begin{figure*}[htbp]
\centerline{\includegraphics{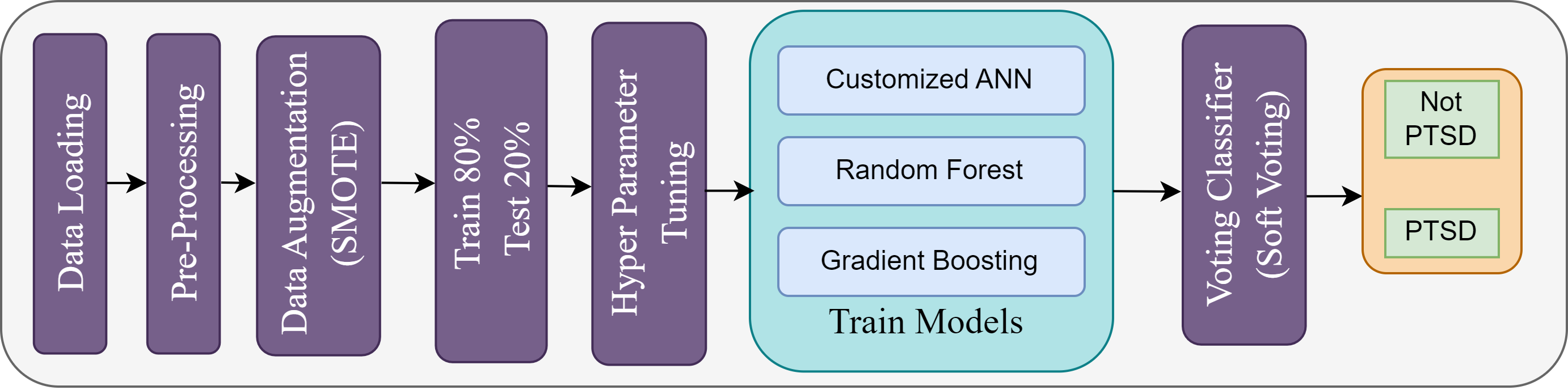}}
\caption{Training and Validation Accuracy.}
\label{fig:proposed_model}
\end{figure*}

Figure \ref{fig:proposed_model} illustrates the architecture of the proposed model for classifying PTSD outcomes. The methodology begins with a comprehensive preprocessing pipeline, including data cleaning and handling missing values using SimpleImputer. To address class imbalances in the dataset, Synthetic Minority Oversampling Technique (SMOTE) is applied for data augmentation. The dataset is then split into 80\% for training and 20\% for testing. Three different models are trained in this framework: a customized Artificial Neural Network (ANN), Random Forest, and Gradient Boosting. Each model is individually optimized to enhance its performance. The ANN captures complex patterns and relationships within the data, while Random Forest and Gradient Boosting excel at handling feature interactions and mitigating overfitting. These models are then integrated using a majority voting technique in a Voting Classifier, which averages the predictions from each model (soft voting) to boost accuracy. The proposed ensemble approach leverages the strengths of all models, resulting in more reliable and accurate PTSD outcome predictions. By combining diverse classifiers, the method improves the robustness of the prediction, making it particularly effective in disaster-prone populations where accurate PTSD assessments are critical for timely mental health interventions.

\subsection{Dataset Description}
\label{sec: Dataset Description}
The dataset for this study was derived from surveys administered to individuals affected by natural disasters in Bangladesh. It offers an in-depth analysis of community sentiments in disaster-prone regions. By synthesizing survey responses from 2,000 individuals, a robust dataset of 8,000 responses has been developed. The study aims to shed light on the mental health challenges arising from environmental disasters, highlighting the intricate relationship between economic instability and psychological well-being. This work represents a significant step forward in understanding and addressing mental health issues in vulnerable areas of Bangladesh.
dataset link - \url{https://data.mendeley.com/datasets/78v9wwg6dp/1}

\subsection{Preprocessing}
\label{sec: Preprocessing}
The target variable was whether the respondents experienced Post-Traumatic Stress Disorder (PTSD) following such events. The preprocessing steps included:
\subsubsection{Data Cleaning} Initially, column names were stripped of leading and trailing whitespace to ensure consistency in data handling. The target column was verified for existence, and categorical features were identified, including age, current occupation, type of disaster faced, Whether they had access to safe shelter post-disaster,Observed mental health issues post-disaster,Mental or physical issues stemming from mental distress and safety during the disaster.
\subsubsection{Missing Value Treatment}
To handle missing values in categorical columns, the SimpleImputer from Scikit-learn was employed, utilizing the strategy of replacing missing values with the most frequent value in each column.
\subsubsection{Encoding Categorical Variables} Categorical variables were transformed into numerical format using label encoding. This process is very important for machine learning algorithms, which require numerical inputs. Each unique category was assigned a corresponding integer value to facilitate this transformation.
\subsection{Data Augmentation}  Synthetic Minority Oversampling Technique (SMOTE) was applied to balance the dataset and enhance the training set. This technique generates synthetic samples for the minority class, thereby enriching the training set and improving the model's ability to generalize across different classes\cite{chawla2002smote}.
\subsection{Feature Scaling and Splitting}
\label{sec: Feature Selection}
Features were standardized using StandardScaler, which ensures each feature contributes equally to model performance. We selected seven categorical features to be input into the models. The target variable indicating PTSD was encoded as binary. The dataset is split into 80\% for training and 20\% for testing.
\subsection{Model Selection and Implementation}
\label{sec: Model Development}
In our analysis, we implemented several models for comparison. First, we employed Logistic Regression, which serves as a baseline model for binary classification. Next, we used the Support Vector Machine (SVM), a powerful classification method particularly effective in high-dimensional spaces. We also incorporated Random Forest, an ensemble method that combines multiple decision trees to enhance accuracy. Additionally, we utilized XGBoost, an optimized gradient boosting framework recognized for its speed and performance. Lastly, we implemented LightGBM, a gradient-boosting framework that employs tree-based learning algorithms specifically optimized for handling large datasets. Each of these models offers unique advantages, allowing for a comprehensive evaluation of their effectiveness in our study.
\subsubsection{Customized ANN Model}
A customized ANN model was constructed using TensorFlow and Keras, designed to classify PTSD occurrence. We constructed a customized ANN model comprising multiple layers with batch normalization and dropout for regularization. The architecture included:
Input Layer: 1024 neurons with ReLU activation.
Hidden Layers: Additional layers with decreasing neuron counts (512, 256, and 128), each followed by Batch Normalization and Dropout for regularization.
Output Layer: A single neuron with sigmoid activation for binary classification. Keras Tuner was utilized to optimize the hyperparameters of the ANN model, including the number of units in dense layers, dropout rates, and learning rates \cite{gulli2017deep}. Hyperparameter tuning is employed for enhancing model accuracy and ensuring that the model generalizes well to unseen data.
\subsubsection{Ensemble Model}
Our ensemble model is designed to leverage the strengths of multiple individual classifiers, enhancing predictive accuracy and robustness in classifying PTSD outcomes. The ensemble approach integrates the predictions from a customized Artificial Neural Network (ANN), Random Forest, and Gradient Boosting showed in figure\ref{fig:proposed_model}. By doing so, it capitalizes on the unique advantages of each model: the ANN excels at capturing complex relationships within the data, while Random Forest provides resilience against overfitting through its ensemble of decision trees. Gradient Boosting further refines the predictions by focusing on minimizing errors iteratively.
The integration is executed through a Voting Classifier that employs soft voting, aggregating the predicted probabilities from each individual model\cite{saha2013combining}. This technique boosted overall predictive performance and facilitated a deeper understanding of the underlying data patterns. 

\section{Result and Discussion} 
\label{sec: Evaluation Metrics}
The effectiveness of each model was assessed through multiple evaluation metrics. These included accuracy, precision, recall, and F1 score, which provided a comprehensive view of their performance. Table \ref{tab:model_comparison} summarizes the performance metrics of various machine learning models used to predict PTSD based on four key performance metrics: accuracy, precision, recall, and F1-score. Among the models evaluated, the ensemble model exhibits the highest accuracy of 96.76\%, demonstrating its effectiveness in classifying instances correctly. The Customized Artificial Neural Network (ANN) follows closely with an accuracy of 94\%, indicating a strong performance as well. In contrast, traditional models like Logistic Regression achieve an accuracy of 84\%, while support vector machines (SVM), Random Forest, XGBoost, and LightGBM all maintain a consistent accuracy of 87\% to 88\%.\\
Precision values reveal that the Ensemble Model and the Customized ANN excel in identifying positive instances, both achieving a precision of 0.97 and 0.96, respectively. This indicates that when these models predict a positive outcome, they are highly accurate in their predictions. The recall metric, which reflects the models' ability to capture all actual positive instances, shows that the Ensemble Model and Customized ANN perform well, with recall values of 0.97 and 0.82, respectively. The F1-scores, which balance precision and recall, further highlight the Ensemble Model's superior performance at 0.96, whereas the Customized ANN shows a lower score of 0.87. Overall, the table illustrates the Ensemble Model as the most robust performer across all metrics, while the other models, particularly the Customized ANN, demonstrate competitive effectiveness, especially in precision and recall, making them valuable choices for classification tasks.
\begin{table}[h]
    \centering
    \caption{Comparison of Model Performance Metrics}
    \label{tab:model_comparison}
    \begin{tabular}{ |l|p{1cm} |p{1cm}| p{1cm}| p{1cm}|}
        \hline
        \textbf{Model} & \textbf{Accuracy (\%)} & \textbf{Precision (\%)} & \textbf{Recall (\%)} & \textbf{F1-Score (\%)} \\
        \hline 
        Logistic Regression & 84.00 & 70.70 & 83.90 & 84.20 \\ 
        SVM & 87.00 & 87.20 & 86.40 & 83.70 \\  
        Random Forest & 87.00 & 86.10 & 87.40 & 86.30 \\  
        XGBoost & 87.00 & 85.70 & 86.80 & 85.80 \\
        LightGBM & 88.00 & 86.40 & 87.80 & 86.80 \\ 
        Customized ANN & 94.00 & 96.00 & 82.00 & 87.00 \\
        Ensemble Model & \textbf{96.76} & \textbf{97.00} & \textbf{97.00} & \textbf{96.00} \\
        \hline
    \end{tabular}
\end{table}

\label{sec: Result and Discussion}
Each model was trained with the same train-test split, utilizing appropriate hyperparameters and optimizations. The ANN model  and the ensamble model employed EarlyStopping and ReduceLROnPlateau for efficient training.
\subsection{Accuracy and Loss Curves}
\label{sec: Accuracy and Loss Curves}
\begin{figure}[htp]
\includegraphics [width= 8 cm]{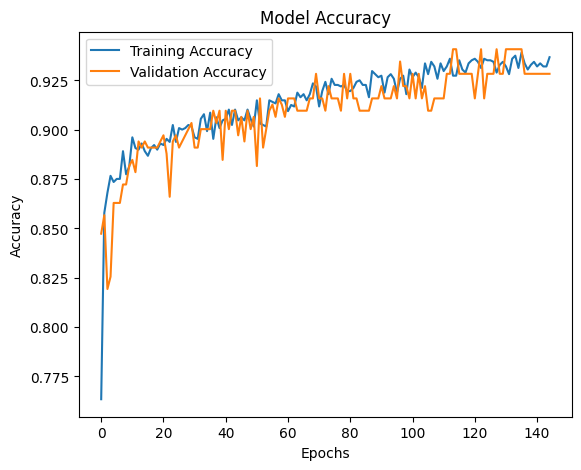}
\caption{Training and Validation Accuracy.}
\label{fig:Accuracy_curve}
\end{figure}

Figure \ref{fig:Accuracy_curve} shows the training and validation accuracy of the ensemble model across epochs. We see that both training and validation accuracy start low but improve steadily over time, converging at around 90-97\%. This steady increase in accuracy suggests that the model is successfully learning from the data. The validation accuracy, which closely follows the training accuracy with minimal fluctuations, indicates that the model is generalizing well to unseen data and is not overfitting. 

\begin{figure}[htp]
\includegraphics[width= 8 cm]{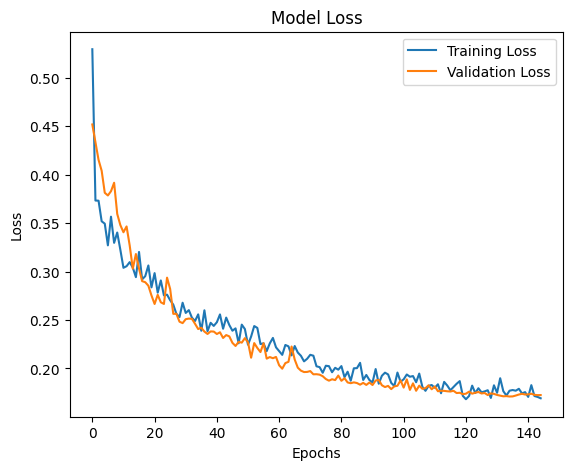}
\caption{Training and Validation Loss.}
\label{fig:Loss_Curve}
\end{figure}

Figure \ref{fig:Loss_Curve} presents the training and validation loss of the ensemble model over the same epochs. The training and validation loss decrease together over the epochs, both stabilizing at low values. The rapid drop in training loss within the first few epochs shows that the model quickly learns patterns within the data, and the low final values suggest effective error minimization. 
Overall, these curves suggest a well-tuned model with strong performance across training and validation datasets, likely due to appropriate model architecture and regularization settings.
\subsection{Confusion Matrix for the Best Model}
\label{sec: Confusion Matrix for the Best Model}
Confusion matrices were generated for each model to visualize the performance in terms of true and false positives and negatives, allowing for a more comprehensive analysis of classification outcomes. 
\begin{figure}[htp]
\includegraphics[width= 8 cm]{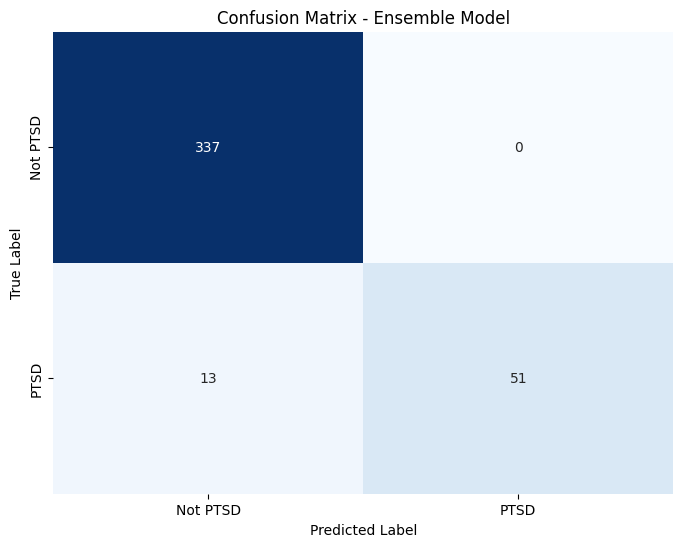}
\caption{Confusion matrix for Ensemble Model.}
\label{fig:Confusion Matrix}
\end{figure}

\subsection{Discussion of Results}
\label{sec: Discussion of Results}
The ensemble model, which combined a customized Artificial Neural Network (ANN), Random Forest, and Gradient Boosting classifiers, achieved an impressive accuracy of 96.76\% in distinguishing between individuals with and without Post-Traumatic Stress Disorder (PTSD). This high accuracy underscores the benefits of employing an ensemble approach to enhance predictive performance.\\
The classification report indicates a precision of 0.96 for class 0 (Not PTSD) and 1.00 for class 1 (PTSD), demonstrating the model's proficiency in accurately detecting PTSD cases. However, the recall for class 1 was 0.80, meaning that 20\% of PTSD cases were incorrectly classified as Not PTSD. The F1-scores of 0.98 for Not PTSD and 0.89 for PTSD reflect a solid balance between precision and recall, albeit with a slight decrease in sensitivity for identifying PTSD cases.\\ 
The confusion matrix revealed that out of 401 predictions, the model correctly identified 337 true negatives and 51 true positives while misclassifying 13 PTSD cases as Not PTSD. This indicates a need for further improvement in detecting PTSD to reduce false negatives.
\begin{figure}[htp]
\includegraphics[width= 8 cm]{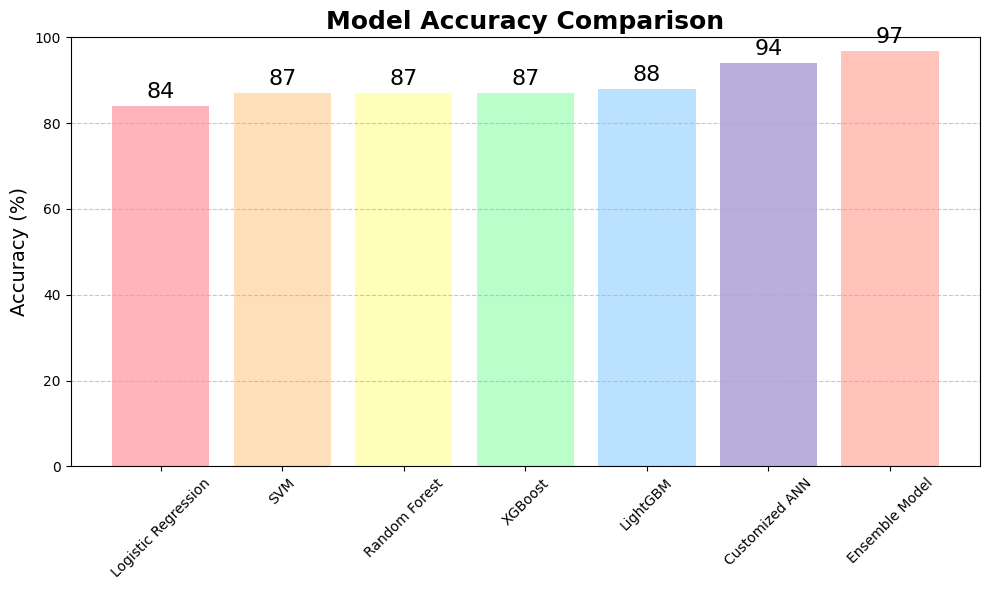}
\caption{Accuracy Comparison of Models.}
\label{fig:Accuracy Comparison of Models}
\end{figure}
Figure \ref{fig:Accuracy Comparison of Models} illustrates the accuracy of all models evaluated, providing a visual representation of each model's performance. This comparison will further highlight the effectiveness of the ensemble approach in achieving the highest accuracy among the tested models. This comprehensive analysis emphasizes the potential for machine learning techniques to aid in the early detection and intervention of PTSD, which is essential for improving mental health outcomes. \\
In summary, while the ensemble model demonstrates strong performance, enhancing sensitivity for PTSD detection remains crucial for future work. Further hyperparameter tuning and feature exploration could lead to improved recall, ensuring better mental health assessments.

\section{Conclusion}
\label{sec: Conclusion}
In this study, we developed an ensemble model for predicting post-traumatic stress disorder (PTSD) outcomes in disaster-prone populations using a range of machine learning algorithms. By implementing a robust preprocessing pipeline that included data augmentation through SMOTE, we effectively addressed class imbalances, thereby enhancing model performance. The integration of a customized Artificial Neural Network (ANN), Random Forest, and Gradient Boosting, combined through a majority voting technique, yielded an impressive accuracy of 96.76\% on a benchmark dataset.
The findings underscore the potential of predictive analytics in mental health research, particularly in understanding and addressing the psychological impacts of disasters. This research not only contributes to the existing body of knowledge on PTSD prediction but also provides valuable insights for policymakers and healthcare providers. By facilitating targeted interventions, our model offers a promising approach to improving mental health outcomes in vulnerable populations affected by traumatic events.
\bibliographystyle{plain}
\bibliography{Reference}

\end{document}